\documentclass{article}

\usepackage{PRIMEarxiv}

\usepackage[utf8]{inputenc} 
\usepackage[T1]{fontenc}    
\usepackage{hyperref}       
\usepackage{url}            
\usepackage{booktabs}       
\usepackage{amsfonts}       
\usepackage{nicefrac}       
\usepackage{microtype}      
\usepackage{amsmath}        
\usepackage{lipsum}
\usepackage{multirow} 
\usepackage{fancyhdr}       
\usepackage{graphicx}       
\graphicspath{{media/}}     
\usepackage{authblk}
\title{RUFNet: Query-Guided Support Mask Refinement and Uncertainty Fusion based on Hybrid Mamba for Few-Shot Brain Tumor Segmentation}

\author[1]{Dongyi He}
\author[2]{Xiangkai Wang}
\author[2]{Binbing Xu}
\author[2]{Bin Jiang}
\author[3]{Hongjie Yan}
\author[4]{Weixiang Liu}
\author[1]{Wai Ting Siok}
\author[1,*]{Nizhuan Wang}

\affil[1]{Department of Language Science and Technology, The Hong Kong Polytechnic University, Hong Kong SAR, China}
\affil[2]{School of Artificial Intelligence, Chongqing University of Technology, Chongqing, China}
\affil[3]{Affiliated Lianyungang Hospital of Xuzhou Medical University, Lianyungang, China}
\affil[4]{College of Mechatronics and Control Engineering, Shenzhen University, Shenzhen, China}
\affil[*]{Correspondence: wangnizhuan1120@gmail.com}

\begin{document}
\maketitle

\begin{abstract}
Few-shot brain tumor segmentation remains challenging due to noisy support masks, inter-patient variations between support and query images, and the lack of pixel-wise confidence estimation. This study proposes RUFNet, a Hybrid Mamba-based few-shot framework that combines support mask refinement with uncertainty-aware posterior fusion. To preserve support-query dependencies with manageable cost, RUFNet adopts a Hybrid Mamba interaction backbone with linear complexity. To reduce support-mask noise, an Attention-Guided Mask Refinement module (AGMR) uses query features to recalibrate support masks and improve prototype consistency. To handle ambiguous predictions, an Uncertainty-Aware Posterior Fusion module (UAPF) estimates pixel-wise variance and adaptively balances few-shot predictions with query-aligned priors. On the Brain Tumor Segmentation Challenge (BraTS) 2020 dataset, RUFNet achieves Dice coefficients of 84.3\% and 86.1\% in the 1-way 1-shot and 1-way 5-shot settings, respectively, outperforming the compared state-of-the-art methods. These results suggest that Hybrid Mamba interaction, mask refinement and uncertainty modelling can improve the robustness of few-shot medical image segmentation. The official implementation code is available at \url{https://github.com/hdy6438/RUFNet}.
\end{abstract}

\keywords{Few-shot learning \and Brain tumor segmentation \and Mask refinement \and Uncertainty estimation \and Mamba}

\section{Introduction}\label{introduction}

Brain tumor segmentation in multi-modal magnetic resonance imaging (MRI) is a fundamental task in neuro-oncology image analysis. Accurate delineation of tumor regions and subregions supports diagnosis, treatment planning and longitudinal assessment, but manual annotation is time-consuming and depends on specialist expertise. Deep learning has therefore become an important approach for automated brain tumor segmentation \cite{1}.

Despite some progress, high-performing segmentation networks remain annotation-intensive. nnU-Net adapts preprocessing and architecture to each dataset, but still assumes enough labelled cases for reliable configuration and training \cite{2}. Swin-UNet introduces Transformer-style long-range modelling, but its attention-based design is data hungry and computationally heavier when dense image interactions are required \cite{3}. These requirements are difficult to satisfy in clinical settings, where rare tumor subtypes, small patient cohorts and expensive voxel-level annotation limit labelled data. Few-shot learning can reduce this burden, but brain tumor models remain vulnerable because tumor appearance, size and location vary substantially across patients \cite{4}.

Existing few-shot segmentation methods address parts of this problem, but leave important gaps. PANet uses prototype alignment, yet its prototypes can be corrupted by inaccurate support masks \cite{11}. SENet improves channel recalibration, but does not explicitly model support-query spatial correspondence \cite{12}. SSL-ALPNet and RPNet add semi-supervised or prototype-based adaptation, but support-query distribution shifts can still weaken feature alignment \cite{13,14}. AAS-DCL and SRCL introduce anatomical supervision, contrastive learning or self-reference constraints, while RegFSL uses registration to reduce cross-case variation \cite{15,16,17}. These methods improve few-shot alignment, but mainly rely on local, fixed or registration-based fusion and do not explicitly preserve long-range support priors during query decoding.

Hybrid Mamba is suitable for this setting because selective state-space modelling captures long-range dependencies with linear complexity. Compared with pure attention, this design is better matched to dense medical images where support-query interactions are spatially extended. The Hybrid Mamba Network (HMNet) further introduces a Support Reset Module and Query Isolation Module to preserve support information and reduce interference from query variation \cite{8}. However, Hybrid Mamba alone does not repair noisy support masks or quantify uncertain boundary predictions.

Several technical barriers further limit the reliability of few-shot brain tumor segmentation. First, noisy or imprecise support masks contaminate prototype construction; guidance-type refinement strategies help, but the support mask itself is often treated as fixed \cite{9,11}. Second, distributional differences between support and query images weaken cross-patient alignment, even when semi-supervised, contrastive or registration-based strategies are used \cite{13,14,15,16,17}. Third, uncertainty-guided and probabilistic segmentation studies show the value of confidence modelling, but few-shot predictions are often fused without explicit pixel-wise posterior weighting \cite{5,6,7,10}. These limitations are clinically important because boundary errors or overconfident predictions may reduce trust in automated segmentation.

To address these challenges, this study proposes RUFNet, a few-shot brain tumor segmentation framework that integrates Hybrid Mamba interaction, support mask refinement and uncertainty-driven posterior fusion. First, Hybrid Mamba is used as the support-query backbone to maintain long-range class priors with manageable computational cost. Second, the Attention-Guided Mask Refinement module (AGMR) uses query features to recalibrate support masks, reducing prototype degradation caused by noisy or mismatched support annotations. Third, the Uncertainty-Aware Posterior Fusion module (UAPF) models pixel-wise variance and adaptively combines meta-predictions with query-aligned priors in ambiguous regions \cite{7}. Together, these components improve support information at the representation level and introduce confidence-aware modulation at the inference level, providing a robust and uncertainty-aware framework for few-shot brain tumor segmentation.

\section{Methods}\label{methods}

This section describes RUFNet, a few-shot brain tumor segmentation framework based on support mask refinement and uncertainty-aware fusion. The framework targets three related limitations of existing few-shot segmentation pipelines: degradation of prototype representations caused by support mask noise, cross-domain variation between support and query images, and the absence of pixel-wise confidence estimates for the final prediction.

\begin{figure}[t]
\centering
\includegraphics[width=0.95\textwidth]{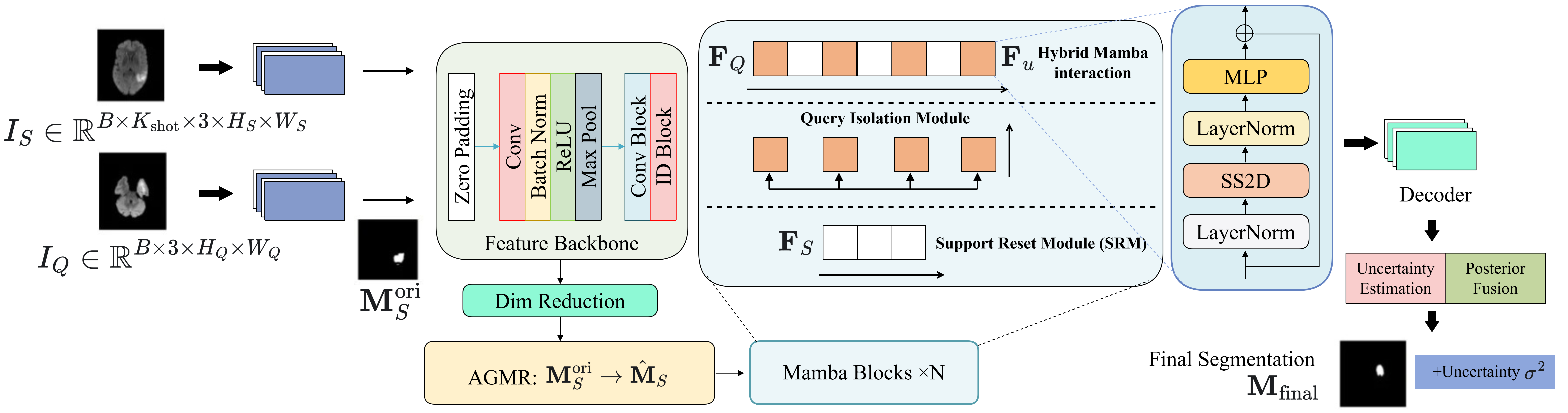}
\caption{Overall architecture of RUFNet. Support and query images are encoded by a shared backbone, the support mask is refined by AGMR, and Hybrid Mamba blocks model support-query interaction. UAPF then fuses the meta-prediction with the refined prior to produce the final segmentation.}
\label{fig:rufnet}
\end{figure}

As shown in Fig.~\ref{fig:rufnet}, RUFNet is organised as an end-to-end inference pipeline with three stages. First, a shared backbone extracts support and query image features and combines them with the initial support mask to form class priors. Second, query-support feature similarity is used to generate attention maps that refine the support mask, improving semantic consistency across patients. Third, a pixel-wise uncertainty branch estimates a variance map and converts it into a fusion weight field that modulates the fusion between meta-predictions and query-aligned prior information, yielding a more stable final segmentation.

\subsection{Hybrid Mamba Feature Backbone}\label{hybrid-mamba-feature-backbone}

RUFNet uses a Hybrid Mamba-based support-query interaction backbone \cite{8} to exploit foreground priors from support samples. Mamba is a selective state-space model that captures long-range dependencies with linear complexity, avoiding the quadratic cost of conventional attention.

The Hybrid Mamba Network (HMNet) models cross-sequence dependencies through a Support Reset Module (SRM) and a Query Isolation Module (QIM). SRM mitigates support feature forgetting by periodically inserting support features into the query sequence, allowing the hidden state to retain class priors throughout the scan. QIM reduces interference from intra-query variation by decoupling inter-pixel dependencies within the query, so that each query pixel draws information primarily from the support state.

The combination of SRM and QIM provides a Hybrid Mamba interaction backbone for fusing class priors across images. This backbone is extended with AGMR and UAPF to address three key failures in few-shot segmentation: amplification of mask noise, cross-domain mismatch and high prediction uncertainty.

\subsection{Attention-Guided Mask Refinement Module (AGMR)}\label{attention-guided-mask-refinement-module}

\begin{figure}[t]
\centering
\includegraphics[width=0.92\textwidth]{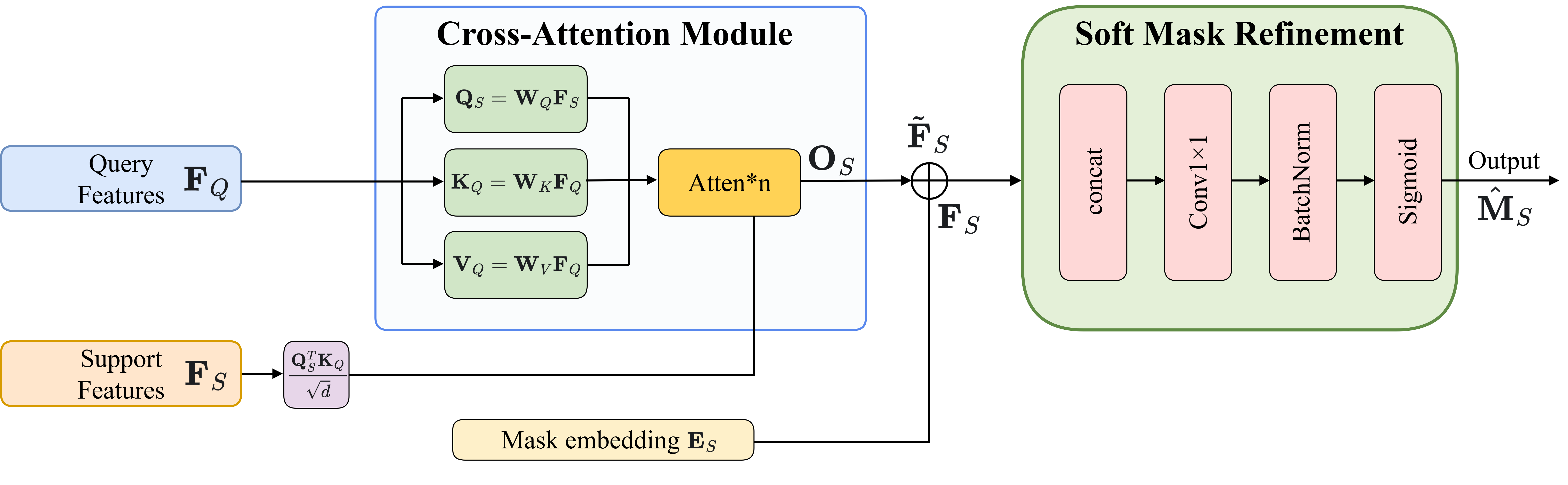}
\caption{Architecture of the Attention-Guided Mask Refinement (AGMR) module. Cross-attention aligns query and support features, and the support-mask embedding is fused with the aligned representation to generate a soft refined mask \(\hat{\mathbf{M}}_S\).}
\label{fig:agmr}
\end{figure}

AGMR takes advantage of distributional information from the query features to refine the support mask. The aim is to align the mask representation more closely with the structural patterns of the query domain, thereby reducing the influence of annotation noise, boundary ambiguity and cross-case variation in the support sample. The module is shown in Fig.~\ref{fig:agmr}.

Given vectorised support features \(\mathbf{F}_{S} \in \mathbb{R}^{C \times N_S}\) and query features \(\mathbf{F}_{Q} \in \mathbb{R}^{C \times N_Q}\), where \(N_S=H_SW_S\) and \(N_Q=H_QW_Q\), pixel-wise similarity is first modelled using cross-attention. Support features are projected to queries \(\mathbf{Q}_S=\mathbf{W}_Q\mathbf{F}_S\), and query features are projected to keys \(\mathbf{K}_Q=\mathbf{W}_K\mathbf{F}_Q\) and values \(\mathbf{V}_Q=\mathbf{W}_V\mathbf{F}_Q\). The projected channel dimension is denoted by \(d\). The support-to-query semantic matching score is computed by scaled dot-product attention:
\begin{equation}
\mathbf{A}_{S \leftarrow Q} =
\operatorname{Softmax}\left(\frac{\mathbf{Q}_S^T\mathbf{K}_Q}{\sqrt{d}}\right)
\in \mathbb{R}^{N_S \times N_Q}.
\end{equation}

The attention matrix re-aggregates query evidence for each support location to generate query-enhanced support features. After channel restoration by \(\mathbf{W}_O\), these features are residually fused with the original support features:
\begin{equation}
\mathbf{O}_S=\mathbf{V}_Q\mathbf{A}_{S \leftarrow Q}^{T},\qquad
\tilde{\mathbf{F}}_S=\mathbf{F}_S+\gamma\mathbf{W}_O\mathbf{O}_S ,
\end{equation}
where \(\gamma\) is a learnable fusion coefficient. The original binary mask \(\mathbf{M}_S^{\mathrm{ori}}\) is embedded into a continuous feature space, \(\mathbf{E}_S=\mathbf{W}_M\mathbf{M}_S^{\mathrm{ori}}\), so that the mask can act as an updatable structural signal during fusion. The refined support features \(\tilde{\mathbf{F}}_S\), original support features \(\mathbf{F}_S\) and mask embedding \(\mathbf{E}_S\) are then concatenated along the channel dimension. Let \(g_{1\times1}(\cdot)\) denote a 1$\times$1 convolution followed by batch normalization (BN) and sigmoid activation. The refined mask is generated as:
\begin{equation}
\begin{aligned}
\mathbf{Z}_S &=
\operatorname{Concat}(\tilde{\mathbf{F}}_S,\mathbf{F}_S,\mathbf{E}_S),\\
\hat{\mathbf{M}}_S &=
\operatorname{Sigmoid}\left(
\operatorname{BN}\left(
\operatorname{Conv}_{1\times1}(\mathbf{Z}_S)
\right)\right).
\end{aligned}
\end{equation}

The resulting continuous mask \(\hat{\mathbf{M}}_S\) preserves intermediate values in noisy and boundary regions, representing ambiguous ``partial foreground'' semantics and providing a probabilistic basis for subsequent uncertainty estimation \cite{9}. By reconstructing the support mask according to query semantics, AGMR aligns the support region more closely with the tumor distribution in the query domain and reduces prototype degradation caused by support noise and cross-domain differences.

\subsection{Uncertainty-Aware Posterior Fusion Module (UAPF)}\label{uncertainty-aware-posterior-fusion-module}

\begin{figure}[t]
\centering
\includegraphics[width=\columnwidth]{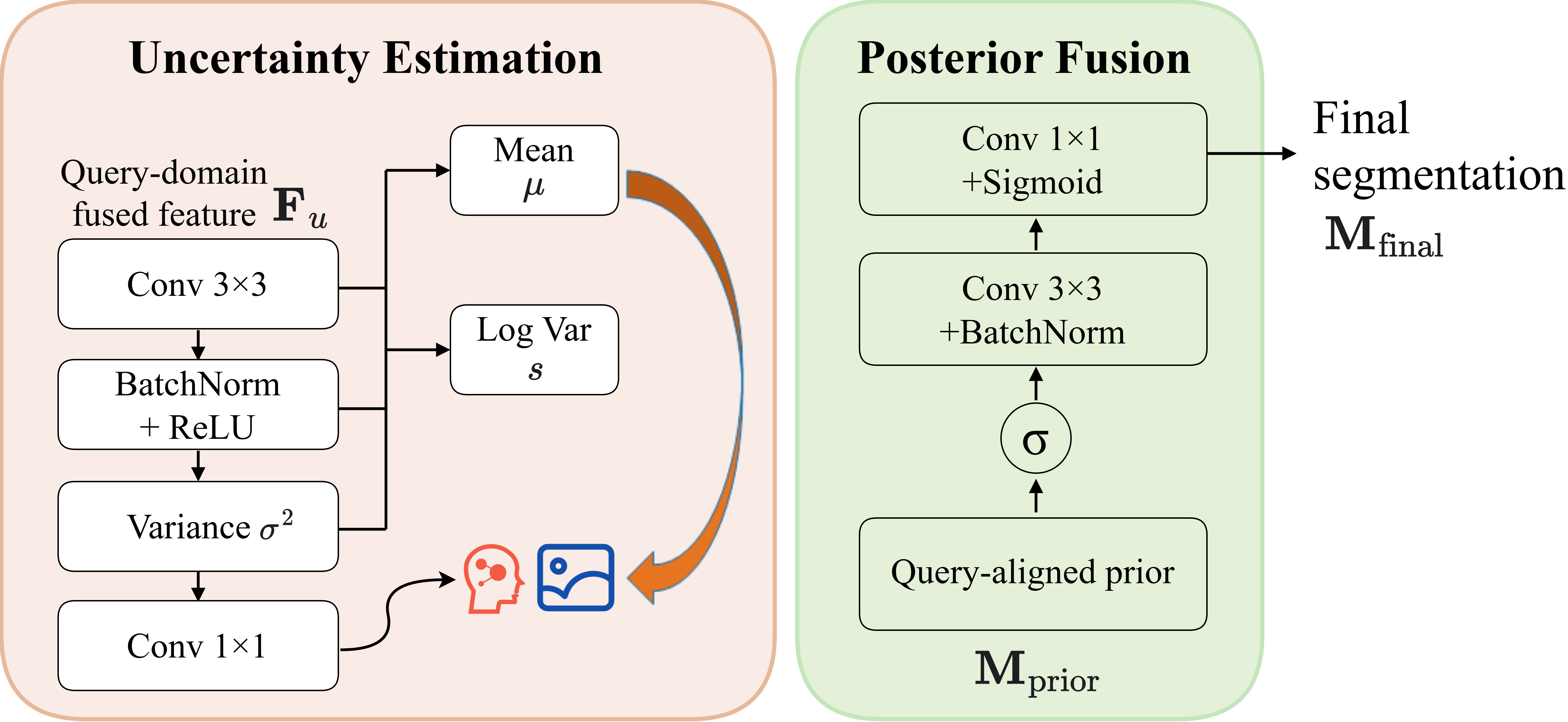}
\caption{Architecture of the Uncertainty-Aware Posterior Fusion (UAPF) module. The query-domain fused feature \(\mathbf{F}_u\) is used to estimate pixel-wise variance, which guides posterior fusion between the meta-prediction \(\mathbf{M}_{\mathrm{meta}}\) and the query-aligned prior \(\mathbf{M}_{\mathrm{prior}}\).}
\label{fig:uapf}
\end{figure}

UAPF explicitly models pixel-wise uncertainty by estimating prediction variance. The resulting uncertainty map controls the spatially adaptive fusion between few-shot predictions and prior information, improving stability in ambiguous regions. The module is shown in Fig.~\ref{fig:uapf}.

Let \(\mathbf{F}_{u}(x)\) denote the query-domain fused feature at pixel \(x\) after Hybrid Mamba interaction and AGMR-guided support prior injection. The refined support mask \(\hat{\mathbf{M}}_S\) is projected to the query domain to form the query-aligned prior \(\mathbf{M}_{\mathrm{prior}}(x)\). The fused feature is mapped to the mean and log-variance of a latent segmentation logit by a feature extractor \(\Phi(\cdot)\) and prediction head \(\Psi(\cdot)\):
\begin{equation}
(\mu(x),s(x))=\Psi(\Phi(\mathbf{F}_{u}(x))),\qquad
\sigma^2(x)=\exp(s(x))
\end{equation}
where \(\mu(x)\) and \(\sigma^{2}(x)\) denote the mean and variance of the latent logit distribution at pixel \(x\), respectively, and \(s(x)\) is the predicted log-variance. The latent logit is modelled as \(z(x)\mid\mathbf{F}_{u}(x)\sim\mathcal{N}(\mu(x),\sigma^2(x))\), and the meta-prediction probability is \(\mathbf{M}_{\mathrm{meta}}(x)=\operatorname{Sigmoid}(\mu(x))\). The variance \(\sigma^2(x)\) is used as an uncertainty measure, with lower values indicating more reliable regions and higher values indicating sparse or ambiguous evidence. The variance is mapped to a weight field that decreases monotonically as uncertainty increases:
\begin{equation}
w(x) = \exp(-\alpha\sigma^2(x))
\end{equation}
where \(\alpha>0\) controls the sensitivity of the weight to variance. When \(\sigma^{2}(x)\) approaches zero, \(w(x)\) $\approx$ 1, indicating high confidence at that location. When \(\sigma^{2}(x)\) is large, \(w(x)\) approaches zero, shifting the prediction towards the query-aligned prior rather than the few-shot output. This weight field is used to perform continuous weighted fusion between the few-shot segmentation output \(\mathbf{M}_{\mathrm{meta}}(x)\) and the prior \(\mathbf{M}_{\mathrm{prior}}(x)\), yielding the posterior prediction:
\begin{equation}
\mathbf{M}_{\mathrm{post}}(x)=
w(x)\mathbf{M}_{\mathrm{meta}}(x)
+(1-w(x))\mathbf{M}_{\mathrm{prior}}(x).
\end{equation}

To enhance spatial coherence, \(\mathbf{M}_{\mathrm{post}}\) is passed through a refinement module consisting of a 3$\times$3 convolution and batch normalization to reduce artefacts from linear interpolation, followed by a 1$\times$1 convolution and sigmoid activation to obtain the final output \(\mathbf{M}_{\mathrm{final}}\). UAPF therefore integrates uncertainty estimation and posterior fusion in a single end-to-end module, allowing the model to identify less reliable regions and adjust the final prediction accordingly.

\section{Datasets and Experimental Setup}\label{datasets-and-experimental-setup}

\subsection{Dataset Processing and Episode Construction}\label{dataset-processing-and-episode-construction}

BraTS 2020 was used as the benchmark dataset. The T1-weighted (T1), gadolinium-enhanced T1-weighted (T1-Gd), T2-weighted (T2) and fluid-attenuated inversion recovery (FLAIR) modalities with voxel-level tumor labels were used. All volumes were processed with the standard BraTS pipeline and resampled to 240$\times$240$\times$155 voxels. Because BraTS 2020 has been widely described in previous studies, this section focuses on the processing steps used for RUFNet.

Given the large variation in tumor volume, morphology and spatial location across BraTS 2020 cases, using all slices directly may amplify support mask noise and cross-case distribution shifts. The labelled samples were therefore screened before constructing few-shot tasks. For each case, the pixel-wise proportion of the tumor region was calculated, and slices with tumor proportions below 0.5\% or above 30\% were excluded. This filtering retained slices with a moderate tumor burden, clearer boundaries and more consistent annotations.

Unlabelled MRI data were also used for unsupervised pre-training. A subset of additional BraTS 2020 cases without publicly released annotations was converted from 3D volumes into 2D transverse slices. Content-based filtering was then applied by estimating the proportion of brain tissue in each slice. Slices dominated by background or containing insufficient tissue information were discarded, and slices with richer brain structures were retained for feature pre-training.

Few-shot episodes were constructed under a 1-way configuration with \(K_{\mathrm{shot}}\in\{1,5\}\) annotated support samples. Support and query samples were partitioned using a patient-level mutual exclusion rule, ensuring that samples in each episode came from different patients. This design reduces support-query information leakage and label-correlation bias. All 2D slice-level samples were processed with the same normalisation and modality enhancement pipeline before being fed into the network. The resulting labelled and unlabelled partitions used for training, validation, testing and pre-training are summarised in Table~\ref{tab:dataset}.

\begin{table}[t]
\caption{Partition of the few-shot segmentation dataset}
\label{tab:dataset}
\centering
\begin{tabular}{lcc}
\toprule
\textbf{Dataset Partition} & \textbf{Cases} & \textbf{Slices} \\
\midrule
Meta-training set & 369 & 5916 \\
Meta-validation set & 50 & 800 \\
Meta-test set & 50 & 800 \\
Unsupervised pre-training set & 100 & 5000 \\
\bottomrule
\end{tabular}
\end{table}

\subsection{Training Details}\label{training-details}

The RUFNet encoder was initialised with weights from a Pyramid Scene Parsing Network (PSPNet) ResNet-50 encoder pre-trained on BraTS. The model was trained end-to-end using stochastic gradient descent with a momentum of 0.9, a batch size of 2 and an initial learning rate of 5$\times$10\textsuperscript{-4}. A polynomial learning-rate decay schedule with an exponent of 0.9 was used. Parameters in the VMamba module were optimised with AdamW at a learning rate of 6$\times$10\textsuperscript{-5}. A weight decay of 1$\times$10\textsuperscript{-4} was applied throughout training. To address class imbalance, the objective combined cross-entropy loss and weighted Dice loss, each with a weight of 1.0. Of the 7396 MRI slices, 5916 slices from 369 cases were used for training. Training was performed for 200 epochs on three NVIDIA A5000 GPUs. During testing, the model with the best validation performance was selected, and foreground classes were represented as binary masks against the background.

\subsection{Evaluation Metrics}\label{evaluation-metrics}

Segmentation performance was evaluated using the Dice similarity coefficient (DSC) and Hausdorff distance (HD). Let \(P\) and \(G\) denote the predicted and ground-truth foreground masks, respectively. DSC measures the spatial overlap between the two masks:
\begin{equation}
\mathrm{DSC}(P,G)=
\frac{2\lvert P\cap G\rvert}{\lvert P\rvert+\lvert G\rvert}.
\end{equation}

HD measures the maximum boundary discrepancy between the predicted and ground-truth masks. The directed boundary distance is defined as:
\begin{equation}
h(A,B)=\sup_{a\in A}\inf_{b\in B}\lVert a-b\rVert_2.
\end{equation}

The symmetric Hausdorff distance is then computed as:
\begin{equation}
\mathrm{HD}(P,G)=\mathbf{M}_S^{\mathrm{ori}}
\max\{h(\partial P,\partial G),h(\partial G,\partial P)\},
\end{equation}
where \(\partial P\) and \(\partial G\) denote the boundaries of the predicted and ground-truth masks. DSC is reported as a percentage, and HD is reported in millimetres. A higher DSC and a lower HD indicate better segmentation performance.

\section{Experimental Results and Discussion}\label{experimental-results-and-discussion}

\subsection{Ablation Study}\label{ablation-study}

To assess the contribution of each component, a Mamba-based few-shot segmentation network was used as the baseline, and AGMR and UAPF were evaluated in both 1-way 1-shot and 1-way 5-shot settings. The ablation results are shown in Table~\ref{tab:ablation}.

In the 1-way 1-shot setting, the baseline achieved a DSC of 82.7\% and an HD of 12.34 mm. Adding AGMR increased the DSC to 83.8\% and reduced the HD to 11.22 mm, suggesting that query-guided support mask refinement suppresses annotation noise and boundary deviations. Adding UAPF alone produced a DSC of 83.1\% and an HD of 11.90 mm, indicating that uncertainty modelling helps the model rely on prior information in low-confidence regions. Combining AGMR and UAPF gave the strongest performance in this setting, with a DSC of 84.3\% and an HD of 10.55 mm.

In the 1-way 5-shot setting, the baseline achieved a DSC of 84.5\% and an HD of 9.14 mm. AGMR increased the DSC to 85.2\% and decreased the HD to 8.45 mm, whereas UAPF achieved a DSC of 85.0\% and an HD of 8.73 mm. The full model further improved the DSC to 86.1\% and reduced the HD to 7.67 mm. These results indicate that AGMR and UAPF provide complementary gains, with AGMR improving cross-image mask consistency and UAPF improving the stability of spatial fusion.

\begin{table}[t]
\caption{Ablation study of RUFNet on the BraTS 2020 dataset}
\label{tab:ablation}
\centering
\begin{tabular}{llcc}
\toprule
\textbf{Setting} & \textbf{Model Variant} & \textbf{DSC(\%)$\uparrow$} & \textbf{HD(mm)$\downarrow$} \\
\midrule
\multirow{4}{*}{\textbf{1-way 1-shot}} & Baseline (no refine/fusion) & 82.7 & 12.34 \\
& + AGMR only & 83.8 & 11.22 \\
& + UAPF only & 83.1 & 11.90 \\
& + AGMR + UAPF & \textbf{84.3} & \textbf{10.55} \\
\midrule
\multirow{4}{*}{\textbf{1-way 5-shot}} & Baseline & 84.5 & 9.14 \\
& + AGMR only & 85.2 & 8.45 \\
& + UAPF only & 85.0 & 8.73 \\
& + AGMR + UAPF & \textbf{86.1} & \textbf{7.67} \\
\bottomrule
\end{tabular}
\end{table}

\subsection{Comparative Performance}\label{comparative-performance}

Table~\ref{tab:comparison} compares RUFNet with representative few-shot semantic segmentation methods under 1-way 1-shot and 1-way 5-shot settings using DSC and HD. Early feature aggregation methods, including PANet and SENet, achieved DSC values of 29.43\% and 36.21\%, respectively, in the 1-shot setting, with HD values above 120 mm and 60 mm. These results suggest that direct cross-image fusion based on raw support masks is sensitive to annotation noise and anatomical variation, leading to unstable foreground prototypes across patients.

Methods with stronger supervised alignment strategies, such as AAS-DCL, SRCL and RegFSL, improved the DSC to 71\%--75\% and reduced the HD to 10--15 mm. This indicates that additional alignment and regularisation can improve discrimination in few-shot segmentation. However, these methods generally use global or fixed fusion strategies, which do not explicitly represent spatially varying prediction risk. As a result, errors may still propagate in regions with ambiguous tumor boundaries.

RUFNet combines Hybrid Mamba-based support-query interaction with AGMR and UAPF. In the 1-way 1-shot setting, it achieved a DSC of 84.3\% $\pm$ 1.2 and an HD of 10.55 mm $\pm$ 2.4, improving the DSC by approximately 9.1 percentage points over RegFSL while keeping the mean HD at a similar level. Although RegFSL obtained a slightly lower mean HD in this setting (10.03 mm), its much larger standard deviation (8.2 mm) indicates considerable episode-level variability. By contrast, the lower HD standard deviation of RUFNet (2.4 mm) suggests more stable boundary localisation under scarce support information. In the 1-way 5-shot setting, RUFNet achieved a DSC of 86.1\% $\pm$ 0.3 and an HD of 7.67 mm $\pm$ 3.5. These findings support the value of query-guided mask refinement for reducing support mask degradation and of pixel-level uncertainty modelling for robust few-shot fusion.

\begin{table}[t]
\caption{Quantitative comparison of segmentation performance between RUFNet and other methods}
\label{tab:comparison}
\centering
\begin{tabular}{lcccc}
\toprule
\multirow{2}{*}{\textbf{Methods}} & \multicolumn{2}{c}{\textbf{1-way 1-shot}} & \multicolumn{2}{c}{\textbf{1-way 5-shot}} \\
\cmidrule(lr){2-3}\cmidrule(lr){4-5}
& \textbf{DSC(\%)$\uparrow$} & \textbf{HD(mm)$\downarrow$} & \textbf{DSC(\%)$\uparrow$} & \textbf{HD(mm)$\downarrow$} \\
\midrule
PANet\cite{11} & 29.43 $\pm$ 1.7 & 121.34 $\pm$ 23.6 & 33.96 $\pm$ 1.6 & 87.14 $\pm$ 13.6 \\
SENet\cite{12} & 36.21 $\pm$ 1.6 & 61.87 $\pm$ 11.5 & 45.27 $\pm$ 1.4 & 53.16 $\pm$ 12.5 \\
SSL-ALPNet\cite{13} & 61.89 $\pm$ 2.3 & 33.13 $\pm$ 9.5 & --- & --- \\
RPNet\cite{14} & 63.79 $\pm$ 1.8 & 28.76 $\pm$ 10.7 & --- & --- \\
AAS-DCL\cite{15} & 71.54 $\pm$ 1.3 & 15.01 $\pm$ 7.8 & 71.87 $\pm$ 0.8 & 14.47 $\pm$ 7.9 \\
SRCL\cite{16} & 74.23 $\pm$ 1.5 & 14.29 $\pm$ 8.8 & 76.14 $\pm$ 1.3 & 9.52 $\pm$ 7.4 \\
RegFSL\cite{17} & 75.18 $\pm$ 1.5 & \textbf{10.03 $\pm$ 8.2} & 77.16 $\pm$ 1.6 & 8.79 $\pm$ 6.7 \\
\textbf{Ours} & \textbf{84.3 $\pm$ 1.2} & 10.55 $\pm$ 2.4 & \textbf{86.1 $\pm$ 0.3} & \textbf{7.67 $\pm$ 3.5} \\
\bottomrule
\end{tabular}
\end{table}

\subsection{Visualisation Results}\label{visualisation-results}

\begin{figure}[t]
\centering
\includegraphics[width=0.60\columnwidth]{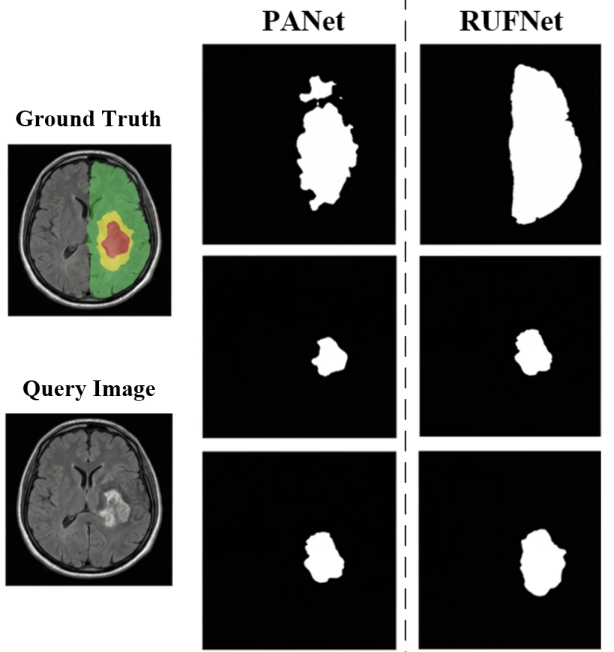}
\caption{Qualitative comparison between PANet and RUFNet under the few-shot setting. RUFNet produces masks that more closely match the ground-truth tumor location and extent.}
\label{fig:qualitative}
\end{figure}

To complement the quantitative evaluation, Fig.~\ref{fig:qualitative} visualises the query image, ground truth and predictions from PANet and RUFNet under the few-shot setting. PANet was selected as a representative prototype-based baseline, and the complete quantitative comparison is provided in Table~\ref{tab:comparison}. With only one support sample, RUFNet delineates tumor regions with contours that are broadly consistent with the ground truth.

\begin{figure}[t]
\centering
\includegraphics[width=\columnwidth]{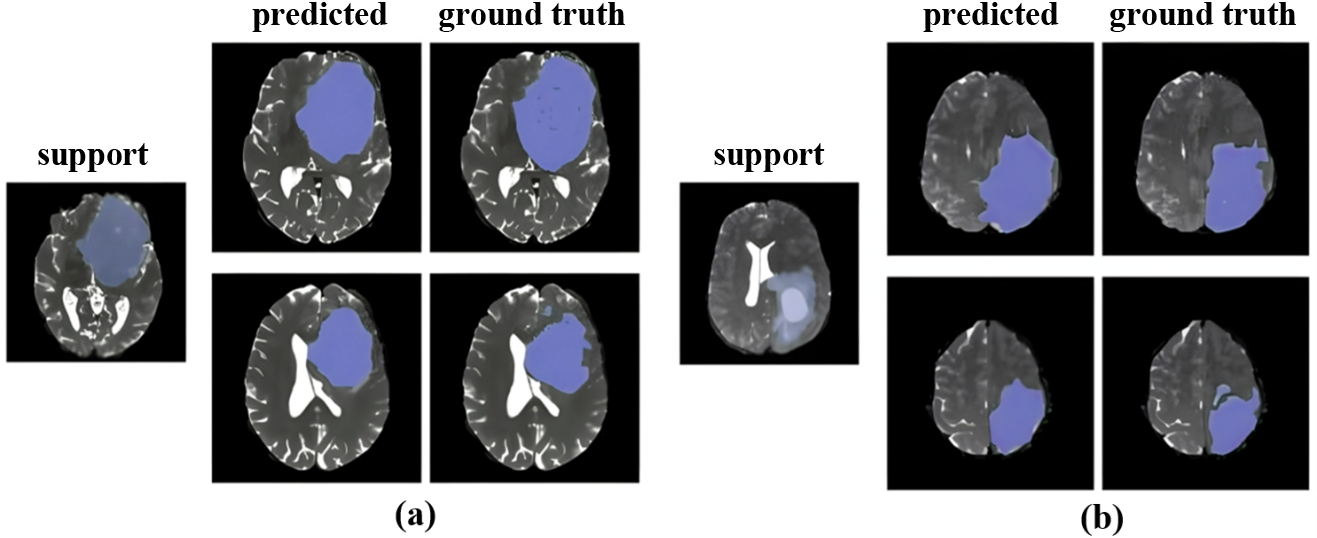}
\caption{Qualitative visualisation of RUFNet under support-set variation. Panels (a) and (b) show representative support-query episodes, with predicted and ground-truth tumor regions highlighted in purple.}
\label{fig:support-generalisation}
\end{figure}

To further examine support-set variation, Fig.~\ref{fig:support-generalisation} visualises predictions for two query MRI slices from different patients, with ground-truth masks as references. Across support samples, the predicted masks remained close to the ground truth in tumor location and boundary shape, indicating stable few-shot segmentation.

These qualitative observations are consistent with the quantitative findings. Query-guided refinement preserved tumor boundaries, while uncertainty-aware fusion reduced unstable predictions in ambiguous regions.

\section{Conclusion}\label{conclusion}

This paper proposed RUFNet, a few-shot brain tumor segmentation framework combining Hybrid Mamba interaction, support mask refinement and uncertainty-aware posterior fusion. The Hybrid Mamba backbone models support-query dependencies, AGMR recalibrates support masks using query semantics, and UAPF performs spatially adaptive fusion through variance modelling. On BraTS 2020, RUFNet achieved DSC scores of 84.3\% and 86.1\% in the 1-shot and 5-shot settings, with HD values of 10.55 mm and 7.67 mm. The ablation and visualisation results indicate that AGMR and UAPF provide complementary gains and support coherent tumor boundary prediction under limited support information. A limitation is that the evaluation used 2D BraTS slices and binary foreground masks without external multi-centre validation. Future work will extend RUFNet to 3D multi-class segmentation and examine uncertainty calibration across scanners and clinical cohorts.

\section*{Acknowledgments}

This research was funded by Natural Science Foundation of Chongqing (CSTB2025NSCQ-JM002, CSTB2025NSCQ-GPX0794, CSTB2024NSCQ-MSX0118), Scientific and Technological Research Program of the Chongqing Education Commission (KJZD-K202303103, KJZD-K202501107, KJQN202501104), Chongqing Municipal Key Project for Technology Innovation and Application Development (CSTB2024TIAD-KPX0042, CSTB2025TIAD-KPX0002), an internal grant from The Hong Kong Polytechnic University (P0048377), The Hong Kong Polytechnic University Departmental Collaborative Research Fund (P0056428), The Hong Kong Polytechnic University Collaborative Research with World-leading Research Groups Fund (P0058097), and Research Grants Council Collaborative Research Fund (C5033-24G).

\bibliographystyle{unsrt}
\bibliography{references}

@article{1,
  title = {Deep learning based brain tumor segmentation: a survey},
  author = {Liu, Z. and Tong, L. and Chen, L. and others},
  journal = {Complex \& Intelligent Systems},
  volume = {9},
  number = {1},
  pages = {1001--1026},
  year = {2023}
}

@article{2,
  title = {nnU-Net: a self-configuring method for deep learning-based biomedical image segmentation},
  author = {Isensee, F. and Jaeger, P. F. and Kohl, S. A. A. and others},
  journal = {Nature Methods},
  volume = {18},
  number = {2},
  pages = {203--211},
  year = {2021}
}

@inproceedings{3,
  title = {Swin-Unet: Unet-like pure transformer for medical image segmentation},
  author = {Cao, H. and Wang, Y. and Chen, J. and others},
  booktitle = {European Conference on Computer Vision},
  pages = {205--218},
  publisher = {Springer Nature Switzerland},
  year = {2022}
}

@article{4,
  title = {Few-shot domain-adaptive anomaly detection for cross-site brain images},
  author = {Su, J. and Shen, H. and Peng, L. and others},
  journal = {IEEE Transactions on Pattern Analysis and Machine Intelligence},
  volume = {46},
  number = {3},
  pages = {1819--1835},
  year = {2021}
}

@article{5,
  title = {From black box AI to XAI in neuro-oncology: a survey on MRI-based tumor detection},
  author = {Asmita and Mittal, P.},
  journal = {Discover Artificial Intelligence},
  volume = {5},
  number = {1},
  pages = {30},
  year = {2025}
}

@article{6,
  title = {Uncertainty-guided Prototype Reliability Enhancement Network for Few-Shot Medical Image Segmentation},
  author = {Hu, J. and Zhou, T. and Huang, K. and others},
  journal = {IEEE Transactions on Medical Imaging},
  volume = {45},
  number = {3},
  pages = {1279--1290},
  year = {2025}
}

@article{7,
  title = {{PULASki}: Learning inter-rater variability using statistical distances to improve probabilistic segmentation},
  author = {Chatterjee, S. and Gaidzik, F. and Sciarra, A. and others},
  journal = {Medical Image Analysis},
  volume = {103},
  pages = {103623},
  year = {2025}
}

@inproceedings{8,
  title = {Hybrid mamba for few-shot segmentation},
  author = {Xu, Q. and Liu, X. and Zhu, L. and others},
  booktitle = {Advances in Neural Information Processing Systems},
  volume = {37},
  pages = {73858--73883},
  year = {2024}
}

@article{9,
  title = {Beyond mask: Rethinking guidance types in few-shot segmentation},
  author = {Chang, S. and Pang, Y. and Zhao, X. and others},
  journal = {Pattern Recognition},
  volume = {165},
  pages = {111635},
  year = {2025}
}

@article{10,
  title = {Uncertainty estimation using a 3D probabilistic U-Net for segmentation with small radiotherapy clinical trial datasets},
  author = {Chlap, P. and Min, H. and Dowling, J. and others},
  journal = {Computerized Medical Imaging and Graphics},
  volume = {116},
  pages = {102403},
  year = {2024}
}

@inproceedings{11,
  title = {PANet: Few-shot image semantic segmentation with prototype alignment},
  author = {Wang, K. and Liew, J. H. and Zou, Y. and others},
  booktitle = {Proceedings of the IEEE/CVF International Conference on Computer Vision},
  pages = {9197--9206},
  year = {2019}
}

@inproceedings{12,
  title = {Squeeze-and-excitation networks},
  author = {Hu, J. and Shen, L. and Sun, G.},
  booktitle = {Proceedings of the IEEE Conference on Computer Vision and Pattern Recognition},
  pages = {7132--7141},
  year = {2018}
}

@inproceedings{13,
  title = {PoissonSeg: semi-supervised few-shot medical image segmentation via poisson learning},
  author = {Shen, X. and Zhang, G. and Lai, H. and others},
  booktitle = {2021 IEEE International Conference on Bioinformatics and Biomedicine},
  pages = {1513--1518},
  publisher = {IEEE},
  year = {2021}
}

@article{14,
  title = {Interactive prototype learning and self-learning for few-shot medical image segmentation},
  author = {Song, Y. and Xu, C. and Wang, B. and others},
  journal = {Artificial Intelligence in Medicine},
  volume = {167},
  pages = {103183},
  year = {2025}
}

@inproceedings{15,
  title = {Dual contrastive learning with anatomical auxiliary supervision for few-shot medical image segmentation},
  author = {Wu, H. and Xiao, F. and Liang, C.},
  booktitle = {European Conference on Computer Vision},
  pages = {417--434},
  publisher = {Springer Nature Switzerland},
  year = {2022}
}

@inproceedings{16,
  title = {Few-shot medical image segmentation regularized with self-reference and contrastive learning},
  author = {Wang, R. and Zhou, Q. and Zheng, G.},
  booktitle = {International Conference on Medical Image Computing and Computer-Assisted Intervention},
  pages = {514--523},
  publisher = {Springer Nature Switzerland},
  year = {2022}
}

@article{17,
  title = {RegFSL: A registration-based framework for few-shot segmentation of brain tumor},
  author = {Farooq, A. and Mishra, D. and Chaudhury, S.},
  journal = {Biomedical Signal Processing and Control},
  volume = {120},
  pages = {110213},
  year = {2026}
}

\end{document}